\documentclass[journal]{IEEEtran}  % For two columns
\ifCLASSINFOpdf
\else
\fi
\usepackage{amsmath,graphicx,cite}
\usepackage{subfig}

\IEEEoverridecommandlockouts
\usepackage{url}
\usepackage{multirow}
\usepackage{hyperref} % byme
\usepackage{nth} % byme
\usepackage{cleveref} % byme
\usepackage{enumerate} % byme
\usepackage{color}
\newcommand{\fig}{Fig.}

\usepackage{algcompatible}

\usepackage[ruled]{algorithm2e}
\usepackage{algpseudocode}
\makeatletter
\renewcommand{\@algocf@capt@plain}{above}% formerly {bottom}
\makeatother

\begin{document}

%\title{N-type PHD Filter for Multiple Target, Multiple Type Filtering}
\title{MRI-PET Registration with Automated Algorithm in Pre-clinical Studies}

\author{Nathanael~L.~Baisa,~\IEEEmembership{Student Member,~IEEE,}
        St{\'e}phanie~Bricq,
        and~Alain~Lalande% <-this % stops a space
       % \\
      %Department of Electrical, Electronic and Computer Engineering \\
      %School of Engineering and Physical Sciences, \\
      %Heriot Watt University, Edinburgh, United Kingdom, \\
      %E-mail: \{nb30, a.m.wallace\}@hw.ac.uk %\vspace{-5mm}
      %E-mail: \{nb30, a.m.wallace, d.e.clark\}@hw.ac.uk %\vspace{-5mm}
\thanks{N. L. Baisa is with the Department of Electrical, Electronic and Computer Engineering, Heriot Watt University, Edinburgh EH14 4AS, United Kingdom. (e-mail: nb30@hw.ac.uk). S. Bricq and A. Lalande are with Burgundy University, Dijon 21078, France.}}% <-this % stops a space
\maketitle

\begin{abstract}

Magnetic Resonance Imaging (MRI) and Positron Emission Tomography (PET) automatic 3-D registration is implemented and validated for small animal image volumes so that the high-resolution anatomical MRI information can be fused with the low spatial resolution of functional PET information for the localization of lesion that is currently in high demand in the study of tumor of cancer (oncology) and its corresponding preparation of pharmaceutical drugs. Though many registration algorithms are developed and applied on human brain volumes, these methods may not be as efficient on small animal datasets due to lack of intensity information and often the high anisotropy in voxel dimensions. Therefore, a fully automatic registration algorithm which can register not only assumably rigid small animal volumes such as brain but also deformable organs such as kidney, cardiac and chest is developed using a combination of global affine and local B-spline transformation models in which mutual information is used as a similarity criterion. The global affine registration uses a multi-resolution pyramid on image volumes of 3 levels whereas in local B-spline registration, a multi-resolution scheme is applied on the B-spline grid  of 2 levels on the finest resolution of the image volumes in which only the transform itself is affected rather than the image volumes. Since mutual information lacks sufficient spatial information, PCA is used to inject it by estimating initial translation and rotation parameters. It is computationally efficient since it is implemented using C++ and ITK library, and is qualitatively and quantitatively shown that this PCA-initialized global registration followed by local registration is in close agreement with expert manual registration and outperforms the one without PCA initialization tested on small animal brain and kidney.

\end{abstract}


% \fi
\IEEEpeerreviewmaketitle

\section{INTRODUCTION} \label{sect:intro}

%Image registration is the process of finding the geometric transformation that maps points from one image to another. It can be grouped into two depending on the similarity criteria: feature-based (also called landmark-based or geometric) and intensity-based (also called area or volume-based or dense or iconic) registration. A feature-based registration method requires extraction of features or salient points common in both images or volumes, and the transformation is estimated from these features by minimizing the distance between them while intensity-based registration optimizes a function measuring the similarity of all geometrically corresponding pixel or voxel pairs, and then obtain the transformation between the entire intensity images. \\

Registration of multi-modality images is the fundamental task in 3-D medical image analysis because it is an instrumental tool in clinical diagnosis and therapy planning by integrating the complementary information of multi-modal images. The common medical imaging modalities are generally divided into two groups depending on the type of information they capture from patients: anatomical imaging modalities and functional imaging modalities. Conventional radiology (X-ray radiography), Computed Tomography (CT), Magnetic Resonance Imaging (MRI) and Ultrasound (echography) provide high resolution images describing their anatomical structure or morphology whereas Positron Emission Tomography (PET), Single Photon Emission Computed Tomography (SPECT) and functional Magnetic Resonance Imaging (fMRI) provide low-resolution functional images for studying the functionality and metabolism of the underlying anatomical structures. Accordingly, images from MRI and PET have entirely different contrast and information content. PET images deliver quantitative data on tumour metabolism containing little anatomic information. In contrast, MRI images provide details of anatomic structures. Therefore, fusing MRI and PET will provide important information on the anatomy-function relationship for the localization of structures or lesions. However, since these two imaging modalities have different resolution, contrast as well as information content, registering them automatically is not trivial~\cite{Cizek2004}.   \\ %~\citet{Cizek2004}

The remainder of the paper is organised as follows: Section~\ref{sect:literature} presents literature review on MRI-PET registration. In section~\ref{sect:design_implement}, the proposed method is discussed, and section~\ref{sect:validation_results} explains the validation techniques used with some obtained results. Finally, section~\ref{sect:conclusion} provides concluding remarks and some future works.

\section{State-of-the-Art} \label{sect:literature}

Many state-of-the-art literatures discuss MRI-PET registration algorithms about a human brain~\cite{Liu2007, Cizek2004, Vaquero2001, Woods1993}. A good review of registration approaches is given in~\cite{Josein2003, Zitova2003}. However, little is known about the performance of these algorithms when applied to small animal images which are, nowadays, very essential for the study of tumor of cancer (oncology). Registration algorithms which normally work on human images may not be as efficient on small animal images because of two reasons. Firstly, many of human registration algorithms are developed taking into account some specific assumptions between imaging modalities~\cite{Vaquero2001}. Secondly, there is a lack of anatomical details in small animal images and often there is a high anisotropy in voxel dimensions i.e. very fine in-plane resolution but large slice thickness when compared to human images on which many registration algorithms are validated~\cite{Vaquero2001}.  \\

An initial preprocessing such as intensity thresholding followed by morphological operations~\cite{Cizek2004} or other segmentation techniques are sometimes needed for good matching and convergence of the registration~\cite{Liu2007}. The majority of multi-modal registration algorithms in literatures use statistical-based similarity metric called mutual information which is recently the most robust similarity measure~\cite{Josein2003}, though correlation ratio is also used in some papers~\cite{Lau2001}. Many papers use parzen windows for density estimation for the computation of mutual information which gives flexibility of using both gradient-based and non-gradient based optimization algorithms for minimizing the cost function~\cite{Xu2008, Mattes2003}. Others use discrete joint histogram density estimation to compute mutual information in combination with non-gradient based optimization algorithms~\cite{Collignon1995}. The performance of different methods for the joint probability estimation to compute mutual information was explained in~\cite{Likar2001}.  \\

Most MRI-PET registration algorithms in the literatures use multi-resolution approach which not only overcomes noise sensitivity but also speeds up registration. Some of these multi-resolution schemes are compared in~\cite{Pluim2001}. This scheme when applied to human brain often gives good result with pre-masked brain since structures outside the brain may be registered instead of the brain itself, especially in low-resolution levels~\cite{Pascau2009}. In addition, pre-processing techniques such as segmentation, masking, and denoising may lead to convergence problems~\cite{Pascau2009}. Hence, it is appropriate to remove these pre-processing steps as used in~\cite{Michael2012} but~\cite{Michael2012} developed algorithm for registering only rigid organ of small animals such as brain and femoral bone using similarity transform. However, the goal of this work is to align image volumes of various anatomical structures and diseases, and both rigid such as brain and deformable image volumes such as kidney, chest, abdomen and heart without taking any account of the type of organ of patient to be captured. Therefore, tailoring any assumption is not appropriate here to pre-process the dataset to be registered.  \\

\section{Algorithm Design and Implementation} \label{sect:design_implement}

In this section, the proposed method and its implementation using C++ and ITK library is discussed.

\subsection{Algorithm Design} \label{sect:design}

For the registration algorithm to be automatic, intensity-based registration should be used rather than feature-based registration since feature-based methods which depend on distinctive points of interest are not suitable for a generic multi-modal registration~\cite{Zitova2003}. This automatic MRI-PET registration requires an energy minimization technique that optimizes a similarity metric (cost function) between these two volumes.  The general energy minimization equation for finding the optimal transformation parameters is given below using MRI as fixed and PET as moving volumes:

\begin{equation}
\centering
   \mathbf{T_g}^*(.)  = \arg\max_{\mathbf{T_g}(.)} \mathbf{S}\big( MRI, \mathbf{T_g}(PET) \big)
\label{eqn:MRI_PET_registration}
\end{equation}

\noindent where $\mathbf{S}$ is the similarity measure (in this case mutual information) between MRI and PET volumes. And $\mathbf{g}$ contains all transformation parameters, $\mathbf{T_g}$ is the intermediate transformation at a certain iteration using transformation parameters in $\mathbf{g}$, while $\mathbf{T_g^*}$ is the optimal transformation that the optimization algorithm is looking for. \\

\begin{figure}
\centering
\includegraphics[width=0.4\textwidth]{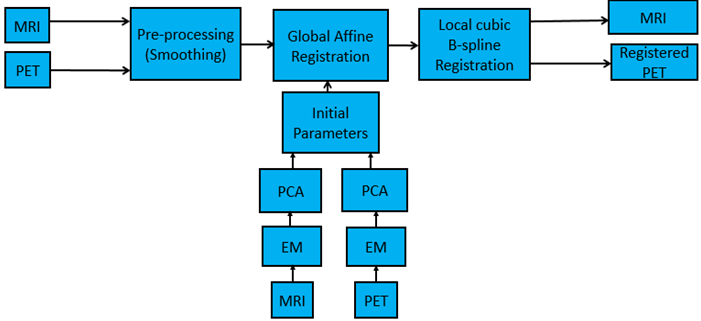}
\caption{The overall developed algorithm in block diagram.}
\label{fig:design}
\end{figure}

The overall developed algorithm is shown in~\fig~\ref{fig:design} using block diagram. First, the fixed MRI volume and the moving PET volume are given as input to the expected maximization algorithm which clusters the volumes by removing the background noise. Then 3-D spatial positions defining the volumes are selected from this clustered volumes and then provided to PCA algorithm which determines the initial translation and rotation parameters. After these initial transformation parameters are computed, both fixed MRI and moving PET volumes are provided in combination with the initialization parameters to the global affine registration framework. This global registration is quite enough for slightly deformable organs such as brain, however, deformable organs such as kidney require locally controlled registration methods. Thus, the output of the global registration is given to the local B-spline registration for fully pipelined registration process. It is a fully automatic registration method without any user-interaction. It is essential to note that only gaussian smoothing when downsampling the volumes for the multi-resolution pyramids is used as a pre-processing which in turn allows to make this registration algorithm not to be specific to any image volume of a patient's organ.

\subsection{Implementation} \label{sect:implement}

Two transformation models are used, one after the other, in this proposed registration algorithm: affine transform which has 12 degrees of freedom (DOF) representing 3 DOF translation, 3 DOF rotation, 3 DOF scaling and 3 DOF shearing, and B-spline free-form deformation which is the most computationally efficient local transform even for a large number of control points since its basis functions have a finite support. \\

Mutual information (MI) is used as a similarity metric for this MRI-PET registration for both global and local registration stages. let $H(M)$, $H(P)$ and $H(M,P)$ be the entropy of MRI volume, entropy of PET volume and the joint entropy of MRI and PET volumes, respectively. These entropies can be expressed mathematically as:

\begin{equation}
\centering
   H(M) = - \sum_{i}^N P_{M} (i)  \log{(P_{M} (i))}
\label{eqn:mri_entropy}
\end{equation}
\begin{equation}
\centering
   H(P) = - \sum_{i}^N P_{P} (i)  \log{(P_{P} (i))}
\label{eqn:pet_entropy}
\end{equation}
\begin{equation}
\centering
   H(M, P) = - \sum_{i}^N \sum_{j}^N P_{M,P} (i,j)  \log{(P_{M,P} (i,j))}
\label{eqn:mripet_entropy}
\end{equation}
\noindent And the MI of the two volumes can be given as:

\begin{equation}
\centering
  MI(M,P) =  H(M) + H(P) -  H(M,P)
\label{eqn:mripet_MI}
\end{equation}

\noindent where $N$ is the number of different possible grayscale values that $i$ and $j$ can take, $P_v(i)$ is the marginal probability of occurrence of a grayscale value $i$ in volume $V$ (either MRI or PET) and $P_{M,P} (i,j)$  is the joint probability of the MRI and PET volumes.\\

Since entropy is defined in terms of probabilistic model of the volume data and no direct access to these probabilistic models, marginal and joint probability density distributions of the MRI and PET volume intensity should be estimated from volume data. Accordingly, parzen windowing is used for the estimation of these density distributions which allows to estimate the continuous joint histogram that can help to derive the close form solution for the derivative of the MI cost function so that gradient-based optimization method can be applied in the registration process~\cite{Xu2008}. Mattes mutual information implemented in ITK library is used in this work which uses parzen windowing for its density estimation. Moreover, it uses one spatial sample set for the whole registration process rather than using new samples at every iterations, which in turn results in a much smoother similarity metric (or cost function) and hence allows the use of the more intelligent gradient-based optimizers~\cite{Mattes2003}.  Therefore, two types of optimization algorithms are used for optimizing MI: Regular step gradient-descent which reduces the learning rate if the gradient changes direction to adjust the direction of the optimization towards the optimal goal for the global registration part, and Limited memory Broyden-Fletcher-Goldfarb-Shannon with bounds to optimize the local registration part which is better for high dimensionality of the parameter space and also allows bound constraints on the independent variables. Moreover, B-spline interplation is used for both global and local registration stages. \\

\begin{figure}[!h]
  \centering
  \subfloat[Coarsest resolution MRI T1 ]
  {\label{fig:lowResMRIT1}
  \includegraphics[height=0.14\textwidth]{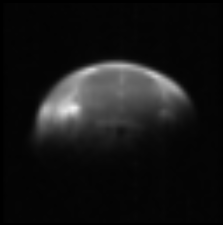}}
  	~~~~~~~~\subfloat[Coarsest resolution PET]
  	{\label{fig:lowResPET}\includegraphics[height=0.14\textwidth]{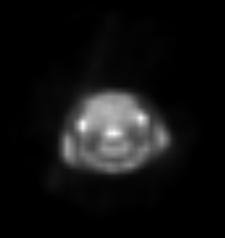}}\\
  \subfloat[MRI T1 labeled by EM]
  {\label{fig:EMmriT1}\includegraphics[height=0.14\textwidth]{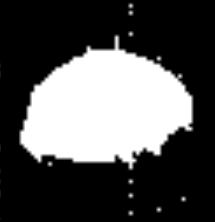}}
  	~~~~~~~~\subfloat[PET labeled by EM]
  	{\label{fig:EMpet}\includegraphics[height=0.14\textwidth]{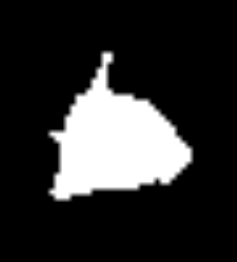}}\\
  \caption{Clustering the image volumes using EM, axial view.}
  \label{fig:clustering}
\end{figure}

Mutual information lacks sufficient spatial information, therefore, it is important to inject it using Principal Component Analysis (PCA) which is used not only for making the registration fully automatic but also for making it robust to local minima by estimating initial parameters. However, before applying PCA, selecting points defining both MRI and PET volumes using clustering techniques is essential. Accordingly, four clustering and/or classification algorithms: expected-maximization (EM), markov random field (MRF), k-means and Bayesian are tested for how efficient they are on these image volumes for removing background noise (or classifying as foreground and background) on the lowest resolution image volumes of a multi-resolution scheme~\cite{Frosio2008}. EM algorithm is more efficient in clustering the required area of image volumes when compared to K-means, MRF and Bayesian and is shown in~\fig~\ref{fig:clustering}.

\noindent And PCA is applied on the 3-D coordinate points (or spatial positions) selected using EM based on image intensities defining the volume and its overall algorithm is given in algorithm 1.

\begin{algorithm}%[H]
\caption{PCA computation for MRI or PET volume}
%\begin{algorithmic}
\SetAlgoLined
\SetKwData{Left}{left}\SetKwData{This}{this}\SetKwData{Up}{up}
\SetKwFunction{Union}{Union}\SetKwFunction{FindCompress}{FindCompress}
\SetKwInOut{Input}{input}\SetKwInOut{Output}{output}
\Input{Volume, Nc = 2: Number of clusters for classifying the volume}
\Output{C, V: 3-D point center of mass C and their principal axes or eigenvectors V}

  \hrulefill \\
  \textit{// Selection~of~points~using~Nc = 2~classes~EM~clustering}\\
%$\textit{Selection~of~points~using~Nc = 2~classes~EM~clustering}$; \\
$P_{3xN} \leftarrow$ 0~Matrix~of~selected~coordinates \\
$Volume \leftarrow$ Nc~classes~EM~clustering  \\
\ForAll{~Voxels(x,y,z) of Volume~}{
  \If{Volume(x,y,z) $>$ 0}
   %{$P \leftarrow Add~Volumes(x,y,z)$;} \\
   {$P \leftarrow Add~coordinates(x,y,z)$;} %\\
}

  \hrulefill \\
  \textit{// Averaging~of~selected~3-D~points~or~coordinates~P}\\
%$\textit{Averaging~of~selected~3-D~points~or~coordinates~P}$; \\
$C \leftarrow \frac{1}{N}\sum_{i=1}^N P_i$~~Centers~of~mass~of~all~points~in~P; \\
\ForAll{~3-D points i of P~}
   {$P_i \leftarrow (P_i - C)$;} %\\

  \hrulefill \\
  \textit{// Eigen-decomposition~(Eigen~analysis)}\\
%$\textit{SVD~decomposition~(Principal~orthogonal~axes)}$; \\
%$\textit{Eigen-decomposition~(Eigen~analysis)}$; \\
$M_{3x3} \leftarrow \frac{1}{N}P.P^T$~~Covariance~matrix; \\
$V,\lambda \leftarrow$ Eigenvectors~and~associated~eigenvalues \\
~of~the~covariance~matrix~M; \\
$V \leftarrow Sort~V~using~\lambda~in~descending~order~and~make $\\
$the~1st~row~of~V~corresponds~to~the~largest~\lambda~and~so~on$;
\end{algorithm}

If $C_P$ and $C_M$ are the centroids of data $PET$ and $MRI$ and $V_P$ and $V_M$ are the eigenvector matrix of $PET$ and $MRI$, then for any point $X_P$, its position $X_M$ in the $MRI$ scan space can be calculated as:

\begin{equation}
X_M = (V_M V_P^T)(X_P - \mu_P) + \mu_M
\label{eqn:X_Mscan}
\end{equation}
\noindent The initial rotation matrix by which the moving volume (PET) should be rotated, and the initial translation vector that is used to translate the PET volume to align it to the fixed MRI volume can be separated and expressed independently as follows:

\begin{eqnarray}
X_M & = & (V_M V_P^T)(X_P - \mu_P) + \mu_M  \nonumber \\
    & = & V_M V_P^T X_P - V_M V_P^T \mu_p + \mu_M \nonumber \\
    & = & R_{PM} X_P - R_{PM} \mu_P + \mu_M  \nonumber \\
    & = & R_{PM} X_P + T_{PM}
\label{eqn:X_MscanSep}
\end{eqnarray}

\noindent where $R_{PM}$ is the initial rotation matrix and is expressed as:
\begin{equation}
R_{PM} = V_M V_P^T
\label{eqn:R_pm}
\end{equation}
\noindent Here it is important to check the initial rotation matrix not to be a reflection matrix (determinant = -1) by confirming its determinant not to be different from 1 using checking matrix, $R_{ch} = [1~0~0;~0~1~0;~0~0~det(R_{PM})$. And then the initial rotation matrix in Eq.~\ref{eqn:R_pm} can be modified as:

\begin{equation}
R_{PMc} = V_M R_{PM}V_P^T
\label{eqn:R_pmc}
\end{equation}

\noindent and the translation vector is:
\begin{equation}
T_{PM} = \mu_M - R_{PMc} \mu_P
\label{eqn:T_pm}
\end{equation}

\noindent The affine transformation components are rotation, scaling, shearing and translation. Therefore, let $\mathbf{M}$ be the combined transformation matrix of rotation ($\mathbf{R}$), scaling ($\mathbf{S}$) and shearing ($\mathbf{H}$) matrices. Therefore, in this work, scaling and shearing matrices are assumed as identity for the initialization of the alignment process whereas the determined rotation matrix from PCA is used for initial rotation matrix part, $R = R_{PMc}$. Moreover, the distance between the centroids of MRI and PET volumes is used as the translation vector as shown in Eq.~\ref{eqn:T_pm} which takes into account the rotation of the centroids of the moving volume (PET).

\begin{equation}
x' = Mx + T_{PM}, where~M = R H S = R_{PMc} H S = R_{PMc}
\label{eqn:M_affine}
\end{equation}

\begin{figure}
\centering
\includegraphics[width=0.4\textwidth]{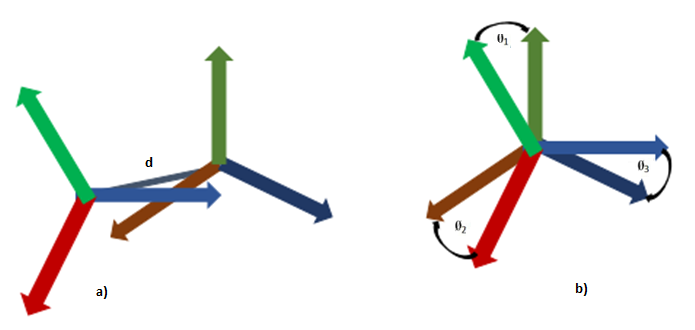}
\caption{Initialization parameters ($d, \theta_1, \theta_2,  \theta_3$) determined from MRI and PET volumes.}
\label{fig:InitialParameters}
\end{figure}

\noindent To illustrate more, Figure~\ref{fig:InitialParameters} shows angles between projected eigenvectors in axial, coronal and sagittal planes from both the MRI and PET volumes and the distance between the center of mass of both volumes: ($d, \theta_1, \theta_2 ~and~ \theta_3)$.

Multi-resolution pyramind of 3 levels is applied on the image volumes in the global affine registration. Note that to generate multi-resolution pyramid image volumes, gaussian smoothing is first performed with the variance set to $(\mathbf{s}/2)^2$, where $\mathbf{s}$ is the shrink factor which is 4 for the lowest resolution, 2 for the middle resolution and 1 for the finest resolution in each dimension. However, in local registration, multi-resolution strategy of 2 levels is applied to the B-spline grid on the finest resolution of MRI and the output of the global registration PET volume. In this multi-resolution scheme, the resolution of the image volume is not modified, but rather the flexibility of the deformable transform itself. \\

In global registration, the number of bins in MRI and PET volumes used to compute the joint histograms is 100 whereas this number is reduced to 50 for the local registration, at all multi-resolution levels for both stages of the registration process. The number of bins is chosen empirically by tuning it for good performance. The number of samples used to estimate the probability density for the global registration is around $9\%$ of the image volume for each multi-resolution level. Besides, for the local registration, 5 number of grids in each dimension is chosen for the low-resolution grid with $7\%$ of the image volume and 6 number of grids in each dimension is selected for the high-resolution grid of B-spline. Since B-spline transform has a large number of parameters, it is important to make sure a large number of samples is selected. Therefore, to make sure larger number of samples is selected compared to the the number of parameters, geometric mean of the total number of pixels in the image volume and the number of parameters used in B-spline transform is used as the number of samples for the high-resolution grid of the  B-spline.  \\

\section{Validation and Results} \label{sect:validation_results}

\subsection{Validation} \label{sect:validation} % experimental data too here
There is no standardized technique that can help to automatically assess the registration accuracy of any medical images so far in the literature.  However, there are techniques which can be used to evaluate the registration algorithms depending on the suitability of the image volumes to be registered: fiducial registration error (FRE), target registration error (TRE), qualitative visual inspection by experts, and interpretation of the similarity measure criterion directly. While FRE and TRE are used to quantitatively evaluate the accuracy of the registration, the other methods qualitatively assess the precision of the registration algorithm~\cite{Fitzpatrick2001}. FRE could not be used here since none of our datasets were acquired with fiducial markers which are artificial markers that could be in place at the time of scan. Therefore, in this MRI-PET multi-modal registration, both quantitative TRE and qualitative visual inspection by experts are used. For the TRE evaluation, a corresponding points (or landmarks) that are available in both MRI and PET image volumes are used though very few landmarks are available since the PET image is very noisy. These same methods are used for MRI-MRI registration too.

\subsection{Results} \label{sect:results}

The performance of this developed algorithm using PCA initialization is evaluated and compared with no PCA initialization hierarchical multi-resolution method. Most registration algorithms so far uses the distance between geometrical centers (centroids) of the volumes as initial translation without removing background noise, and identity matrix for the other parameters of the affine transform. This is referred to as \emph{Without PCA} in the given comparison tables. {\bf VV} 4-D viewer is used for viewing the registered image volumes to assess them qualitatively and quantitatively~\cite{Seroul2008}. The performance of these two registration approaches is tested on four different rats and mice studies which are listed in the Table~\ref{tbl:datasets} with all their information: MRI-PET brain study, MRI-MRI brain study, MRI-PET kidney study, and MRI-MRI kidney study. Accordingly, the PCA initialized registration method outperformes the one without PCA initialization in all study results as shown in their corresponding tables below except that both methods converge successfully on the provided datasets.

\begin{table}[htbp]
\begin{center}
  \begin{tabular}{ |@{}l @{}|@{} c @{}|@{} c @{}|@{} c @{}|@{} r | }
    \hline
    Subject  &  Modality    & Dimension (voxels) & Voxel Size ($mm^3$) & Bits \\ \hline %\hlines
    \multirow{2}{*}{Brain}   &   MRI-T1 and MRI-T2  &  256 x 256 x 8  &   0.156 x 0.156 x 1.8 & 16 \\
          &  FDG-PET          &   175 x 175 x 19  & 0.387 x 0.387 x 0.775   & 16\\ \hline
   \multirow{2}{*}{Brain}   & MRI-T1  &  256 x 256 x 8  &   0.156 x 0.156 x 1.8 & 16 \\
        &  MRI-T2  &  256 x 256 x 8  &   0.156 x 0.156 x 1.8  & 16 \\ \hline
    \multirow{2}{*}{Kidney}  &  MRI-T1 and MRI-T2  &  256 x 256 x 7 & 0.27 x 0.25 x 3.0 & 16 \\
        &   FDG-PET          &   175 x 175 x 24  & 0.387 x 0.387 x 0.775  & 16 \\ \hline
    \multirow{2}{*}{Kidney} &   MRI-T1         &    256 x 256 x 7  &  0.27 x 0.25 x 3.0 & 16 \\
       &   MRI-T2         &    256 x 256 x 7  &  0.27 x 0.25 x 3.0  & 16 \\ \hline
  \end{tabular}
\end{center}
\caption{List of datasets used for registration.}
\label{tbl:datasets}
\end{table}

\subsubsection {MRI-PET brain} \label{sect:mripetbrain}

\begin{figure}[!h]
  \centering
  \subfloat[MRI T1, slice 5]
  {\label{fig:brainT14}
  \includegraphics[height=0.14\textwidth]{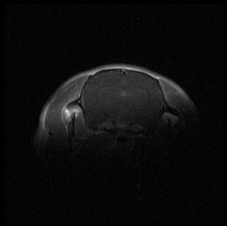}}
  	~~~~~~~~\subfloat[PET, slice 3]
  	{\label{fig:brainPET4}\includegraphics[height=0.14\textwidth]{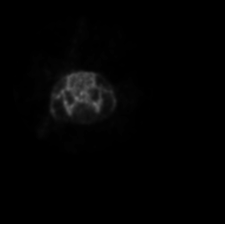}}\\
  \subfloat[Registered using global affine]
  {\label{fig:affinebrainmripet}\includegraphics[height=0.14\textwidth]{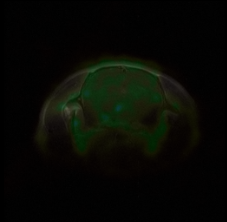}}
  	~~~~~~~~\subfloat[Registered using both affine and local B-spline]
  	{\label{fig:localbrainmripet}\includegraphics[height=0.14\textwidth]{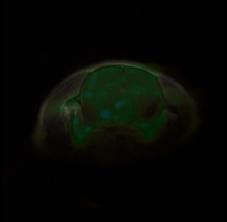}}\\
  \caption{MRI T1 and PET mice brain registration, axial view.}
  \label{fig:brain_mripet}
\end{figure}

MRI T1 and PET volumes are registered and shown in~\fig~\ref{fig:brain_mripet} in axial view. Though the global affine registration does well, correcting the deformation by local registration slightly improves the result and the comparison is given on Table~\ref{tbl:result1}.

\begin{table}[htbp]
\begin{center}
  \begin{tabular}{ |l | c | c | c | c |r | }
    \hline
    Algorithm & TRE (mm) &  Iterations    & Time (sec.) & Quality (MI)   \\ \hline %\hlines
    \multirow{2}{*}{PCA}     & 0.56 (0.38)   & 42 (45) &  36 (33)  &   -0.408 \\
                     &     &       &          &    \\ \hline
    \multirow{2}{*}{Without PCA}   & 1.07 (0.96)    & 46 (47)  &  40 (34)  &  -0.402   \\
               &     &    &     &     \\ \hline
  \end{tabular}
\end{center}
\caption{Mean TRE, iterations and time taken for global and local (in bracket) of rat brain MRI T1-PET images registration.}
\label{tbl:result1}
\end{table}

\subsubsection {MRI-MRI brain}

\begin{figure}[!h]
  \centering
  \subfloat[MRI T2, slice 6 ]
  {\label{fig:brainT13}
  \includegraphics[height=0.14\textwidth]{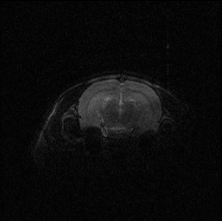}}
  	~~~~~~~~\subfloat[MRI T1, slice 6]
  	{\label{fig:brainT25}\includegraphics[height=0.14\textwidth]{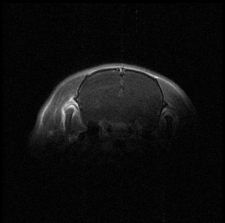}}\\
  \subfloat[Registered using global affine]
  {\label{fig:affinebrainmriT1mriT2}\includegraphics[height=0.14\textwidth]{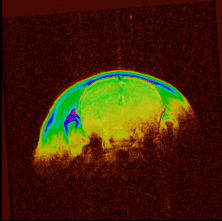}}
  	~~~~~~~~\subfloat[Registered using both affine and local B-spline]
  	{\label{fig:localbrainmriT1mriT2}\includegraphics[height=0.14\textwidth]{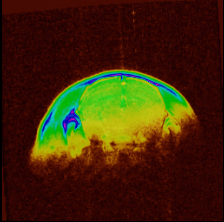}}\\
  \caption{MRI T2 and MRI T1 mice brain registration, axial view.}
  \label{fig:brain_mriT1mriT2}
\end{figure}

The global affine registration works almost perfectly in this MRI T2 (fixed)-MRI T1 (moving) registration as shown in~\fig\ref{fig:brain_mriT1mriT2}. The improvement of the registration result by the addition of the local registration is almost negligible when compared to the multi-modal case as discussed in section~\ref{sect:mripetbrain}. All its quantitative evaluation is shown in Table~\ref{tbl:result2}.

\begin{table}[htbp]
\begin{center}
  \begin{tabular}{ |l | c | c | c| c | r| }
    \hline
    Algorithm & TRE (mm) &  Iterations    &  Time (sec.) & Quality (MI)   \\ \hline %\hlines
    \multirow{2}{*}{PCA}     & 0.07 (0.07)  &  35(42)  &   27 (24) & -0.44  \\
              &   &     &           &    \\ \hline
    \multirow{2}{*}{Without PCA}   & 0.57 (0.42)    &  39(41)   &  37 (29)  &  -0.41   \\
               &    &    &   &     \\ \hline
  \end{tabular}
\end{center}
\caption{Mean TRE, iterations and time taken for global and local (in bracket) of rat brain MRI T1-MRI T2 images registration.}
\label{tbl:result2}
\end{table}

\subsubsection {MRI-PET kidney}
\begin{figure}[!h]\tiny
  \centering
  \subfloat[MRI T1 slice 4]
  {\label{fig:kidneyT14}\includegraphics[width=0.14\textwidth]{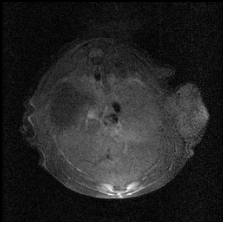}}
  \subfloat[Registered MRI T1 and PET]
  {\label{fig:kidneyregistered4}\includegraphics[width=0.14\textwidth]{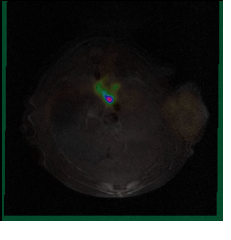}}
  \subfloat[PET slice 12 ]
  {\label{fig:kidneypet12}\includegraphics[width=0.14\textwidth]{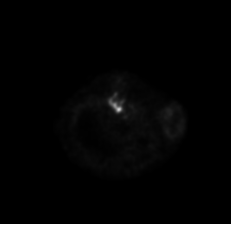}}
   \caption{MRI T1 and PET rat kidney registration, axial view.}
  \label{fig:kidney_mriT1pet}
\end{figure}

Even though there is high anisotropy in pixel dimensions especially in MRI datasets (Table~\ref{tbl:datasets}) and noisy PET image which affect the registration performance, the developed registration algorithm almost overcomes it and the result is shown in~\fig\ref{fig:kidney_mriT1pet}. The addition of local registration stage to the global registration significantly improves the result and its comparison is shown in Table~\ref{tbl:result3}.

\begin{table}[htbp]
\begin{center}
  \begin{tabular}{ |l | c | c | c| c | r| }
    \hline
    Algorithm & TRE (mm) &  Iterations    &  Time (sec.) & Quality (MI)   \\ \hline %\hlines
    \multirow{2}{*}{PCA}     & 0.61 (0.47)   &   47 (44)  &  30 (31)  &  -0.509  \\
              &     &     &          &    \\ \hline
    \multirow{2}{*}{Without PCA}   & 1.23 (1.18)    & 51 (47)  &   33(32.5) &  -0.493   \\
               &     &    &   &     \\ \hline
  \end{tabular}
\end{center}
\caption{Mean TRE, iterations and time taken for global and local (in bracket) of rat kidney MRI T1-PET images registration.}
\label{tbl:result3}
\end{table}

\subsubsection {MRI-MRI kidney}
Registration of monomodal MRI T1 (fixed)-MRI T2 (moving) kidney is much simpler than its multi-modal datasets, hence it takes shorter time or fewer number of iterations as shown in Table~\ref{tbl:result4} when compared to multi-modal registration shown in Table~\ref{tbl:result3}.

\begin{figure}[!h]\tiny
  \centering
  \subfloat[MRI T1 slice 6]
  {\label{fig:kidney2T16}\includegraphics[width=0.14\textwidth]{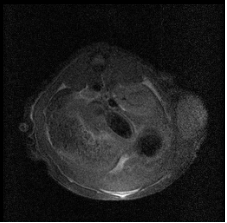}}
  \subfloat[Registered MRI T1 and MRI T2]
  {\label{fig:kidney2registered6}\includegraphics[width=0.14\textwidth]{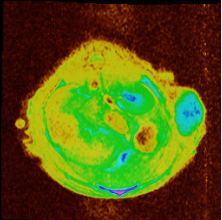}}
  \subfloat[MRI T2 slice 6]
  {\label{fig:kidney2T26}\includegraphics[width=0.14\textwidth]{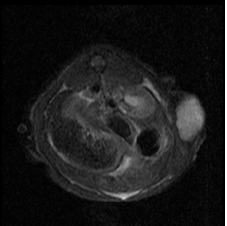}}
  \caption{MRI T1 and MRI T2 rat kidney registration, axial view.}
  \label{fig:kidney_mriT1mriT2}
\end{figure}

\begin{table}[htbp]
\begin{center}
  \begin{tabular}{ |l | c | c | c| c | r| }
    \hline
    Algorithm & TRE (mm) &  Iterations    & Time (sec.) & Quality (MI)   \\ \hline %\hlines
    \multirow{2}{*}{PCA}     & 0.15 (0.09)   &  38 (39)  &   26 (25)  &  -0.736  \\
              &     &      &      &    \\ \hline
    \multirow{2}{*}{Without PCA}   &  0.63 (0.51)    & 48 (42)  &  31 (27)  & -0.654     \\
               &    &    &    &     \\ \hline
  \end{tabular}
\end{center}
\caption{Mean TRE, iterations and time taken for global and local (in bracket) of rat kidney MRI T1-MRI T2 images registration.}
\label{tbl:result4}
\end{table}

\section{Conclusion and Future Works}  \label{sect:conclusion}

In this work, the automated MRI-PET registration in pre-clinical studies is implemented using C++ and ITK library for the better localization of structures or lesion which is very essential in the study of tumor of cancer (oncology) using PCA-initialized global affine registration and the local B-spline registration, one after the other. PCA is applied on 3-D spatial positions which were selected using EM algorithm, and this method is shown to be in close agreement with expert manual registration outperforming the one without PCA initialization tested on small animal brain and kidney.  \\

As a future work, we would like to test the performance of this algorithm on small animal datasets such as cardiac (or heart) and chest which could not be acquired in due of deadline of the submission of this manuscript.

%% use section* for acknowledgment
%\section*{Acknowledgment}
%

\ifCLASSOPTIONcaptionsoff
  \newpage
\fi

\bibliographystyle{IEEEtran}
%\bibliography{strings,refs}
\bibliography{reference}

\begin{thebibliography}{10}
\providecommand{\url}[1]{#1}
\csname url@samestyle\endcsname
\providecommand{\newblock}{\relax}
\providecommand{\bibinfo}[2]{#2}
\providecommand{\BIBentrySTDinterwordspacing}{\spaceskip=0pt\relax}
\providecommand{\BIBentryALTinterwordstretchfactor}{4}
\providecommand{\BIBentryALTinterwordspacing}{\spaceskip=\fontdimen2\font plus
\BIBentryALTinterwordstretchfactor\fontdimen3\font minus
  \fontdimen4\font\relax}
\providecommand{\BIBforeignlanguage}[2]{{%
\expandafter\ifx\csname l@#1\endcsname\relax
\typeout{** WARNING: IEEEtran.bst: No hyphenation pattern has been}%
\typeout{** loaded for the language `#1'. Using the pattern for}%
\typeout{** the default language instead.}%
\else
\language=\csname l@#1\endcsname
\fi
#2}}
\providecommand{\BIBdecl}{\relax}
\BIBdecl

\bibitem{Cizek2004}
J.~Cízek, K.~Herholz, S.~Vollmar, R.~Schrader, J.~Klein, and W.-D. Heiss,
  ``\BIBforeignlanguage{eng}{Fast and robust registration of {PET} and {MR}
  images of human brain},'' \emph{\BIBforeignlanguage{eng}{{NeuroImage}}},
  vol.~22, no.~1, pp. 434--442, May 2004.

\bibitem{Liu2007}
J.~Liu and J.~Tian, ``Registration of brain {MRI/PET} images based on adaptive
  combination of intensity and gradient field mutual information,''
  \emph{International Journal of Biomedical Imaging}, vol. 2007, 2007.

\bibitem{Vaquero2001}
J.~Vaquero, M.~Desco, J.~Pascau, A.~Santos, I.~Lee, J.~Seidel, and M.~Green,
  ``{PET}, {CT}, and {MR} image registration of the rat brain and skull,''
  \emph{{IEEE} Transactions on Nuclear Science}, vol.~48, no.~4, pp.
  1440--1445, 2001.

\bibitem{Woods1993}
R.~P. Woods, J.~C. Mazziotta, and S.~R. Cherry, ``{MRI-PET} registration with
  automated algorithm,'' \emph{Journal of computer assisted tomography},
  vol.~17, no.~4, pp. 536--546, Aug. 1993.

\bibitem{Josein2003}
J.~P.~W. Pluim, J.~Maintz, and M.~Viergever, ``Mutual-information-based
  registration of medical images: a survey,'' \emph{{IEEE} Transactions on
  Medical Imaging}, vol.~22, no.~8, pp. 986--1004, 2003.

\bibitem{Zitova2003}
B.~Zitová and J.~Flusser, ``Image registration methods: a survey,''
  \emph{Image and Vision Computing}, vol.~21, no.~11, pp. 977--1000, Oct. 2003.

\bibitem{Lau2001}
Y.~H. Lau, M.~Braun, and B.~F. Hutton, ``\BIBforeignlanguage{eng}{Non-rigid
  image registration using a median-filtered coarse-to-fine displacement field
  and a symmetric correlation ratio},'' \emph{\BIBforeignlanguage{eng}{Physics
  in medicine and biology}}, vol.~46, no.~4, pp. 1297--1319, Apr. 2001.

\bibitem{Xu2008}
R.~Xu, Y.-W. Chen, S.-Y. Tang, S.~Morikawa, and Y.~Kurumi, ``Parzen-window
  based normalized mutual information for medical image registration,''
  \emph{{IEICE} - Trans. Inf. Syst.}, vol. E91-D, no.~1, p. 132–144, Jan.
  2008.

\bibitem{Mattes2003}
D.~Mattes, D.~R. Haynor, H.~Vesselle, T.~K. Lewellen, and W.~Eubank,
  ``\BIBforeignlanguage{eng}{{PET-CT} image registration in the chest using
  free-form deformations},'' \emph{\BIBforeignlanguage{eng}{{IEEE} transactions
  on medical imaging}}, vol.~22, no.~1, pp. 120--128, Jan. 2003.

\bibitem{Collignon1995}
A.~Collignon, F.~Maes, D.~Delaere, D.~Vandermeulen, P.~Suetens, and G.~Marchal,
  ``Automated multi-modality image registration based on information theory,''
  \emph{Information Processing in Medical Imaging}, pp. 263--274, 1995.

\bibitem{Likar2001}
B.~Likar and F.~Pernuš, ``A hierarchical approach to elastic registration
  based on mutual information,'' \emph{Image and Vision Computing}, vol.~19,
  no. 1–2, pp. 33--44, Jan. 2001.

\bibitem{Pluim2001}
J.~Pluim, J.~Maintz, and M.~Viergever, ``Mutual information matching in
  multiresolution contexts,'' \emph{Image and Vision Computing}, vol.~19, no.
  1–2, pp. 45--52, Jan. 2001.

\bibitem{Pascau2009}
J.~Pascau, J.~D. Gispert, M.~Michaelides, and P.~K. Thanos,
  ``\BIBforeignlanguage{eng}{Automated method for small-animal {PET} image
  registration with intrinsic validation},''
  \emph{\BIBforeignlanguage{eng}{Molecular imaging and biology: {MIB:} the
  official publication of the Academy of Molecular Imaging}}, vol.~11, no.~2,
  pp. 107--113, Apr. 2009.

\bibitem{Michael2012}
B.~Michael, L.~Martin, L.~Roger, T.~Luc, and D.~Maxime, ``New parallel
  multi-resolution approach for an automatic registration of small animal
  {PET}-{MRI},'' Master's thesis, Universit´e de Sherbrooke, Qu´ebec, Canada,
  2012.

\bibitem{Frosio2008}
I.~Frosio, S.~Abati, and N.~A. Borghese, ``\BIBforeignlanguage{en}{An
  expectation maximization approach to impulsive noise removal in digital
  radiography},'' \emph{\BIBforeignlanguage{en}{International Journal of
  Computer Assisted Radiology and Surgery}}, vol.~3, no. 1-2, pp. 91--96, Jun.
  2008.

\bibitem{Fitzpatrick2001}
J.~M. Fitzpatrick and J.~B. West, ``\BIBforeignlanguage{eng}{The distribution
  of target registration error in rigid-body point-based registration},''
  \emph{\BIBforeignlanguage{eng}{{IEEE} transactions on medical imaging}},
  vol.~20, no.~9, pp. 917--927, Sep. 2001.

\bibitem{Seroul2008}
P.~Seroul and D.~Sarrut, ``{VV}: a viewer for the evaluation of {4D} image
  registration,'' 07 2008.

\end{thebibliography}


%\vfill


\end{document}